\documentclass{ifacconf}

\usepackage{graphicx}      % include this line if your document contains figures
\usepackage[authoryear]{natbib}       % required for bibliography
\usepackage{epsfig}        % required to add vector figures
\usepackage{amsmath}
\usepackage{amssymb}
\usepackage{caption}
\usepackage{subcaption}

\begin{document}
\begin{frontmatter}

\title{Unified Manipulability and Compliance Analysis of Modular Soft-Rigid Hybrid Fingers}
%\thanksref{footnoteinfo}} 

%{\thanks}

\author{Jianshu Zhou}, 
\author{Boyuan Liang}, 
\author{Junda Huang}, and
\author{Masayoshi Tomizuka}

\address{Department of Mechanical Engineering, University of California, Berkeley (e-mail: [jianshuzhou@berkeley.edu, liangb@berkeley.edu, ustcdraja@gmail.com, tomizuka@berkeley.edu]).}

\begin{abstract}   
% Abstract of 50--100 words
This paper presents a unified framework to analyze the manipulability and compliance of modular soft-rigid hybrid robotic fingers. The approach applies to both hydraulic and pneumatic actuation systems. A Jacobian-based formulation maps actuator inputs to joint and task-space responses. Hydraulic actuators are modeled under incompressible assumptions, while pneumatic actuators are described using nonlinear pressure–volume relations. The framework enables consistent evaluation of manipulability ellipsoids and compliance matrices across actuation modes. We validate the analysis using two representative hands: DexCo (hydraulic) and Edgy-2 (pneumatic). Results highlight actuation-dependent trade-offs in dexterity and passive stiffness. These findings provide insights for structure-aware design and actuator selection in soft-rigid robotic fingers.
\end{abstract}


\end{frontmatter}
%===============================================================================

\section{Introduction}
\label{sec:intro}

Achieving dexterous and robust manipulation in unstructured environments remains a core challenge in robotics~\citep{Billard2019ScienceManipulation}. As robotic hands advance beyond simple grasping toward versatile in-hand manipulation, there is growing demand for systems that combine precise control, adaptability, compliance, and safe physical interaction. \textit{Soft robotic systems} excel in adaptive grasping through material compliance and morphology-driven control but face limitations in precision tasks. \textit{Soft-rigid hybrid systems}, which integrate compliant actuators with rigid linkages, offer a balanced approach—enabling both fine control and structural adaptability.

A variety of soft and soft-rigid robotic hands has emerged over the past decade. Pneumatic systems like the RBO Hand~\citep{Deimel2016RBOHand} and the Pisa/IIT SoftHand~\citep{Catalano2014SoftHand} enable compliant grasping via soft inflation or tendon-driven underactuation. Hybrid hands enhance control and robustness by combining soft actuators with rigid structures, including origami mechanisms~\citep{Li2019OrigamiSoftRobot}, layer jamming~\citep{Manti2016StiffeningReview}, bellow-type encapsulation~\citep{Lee2024SoftRigidGripper}, and geometric confinement~\citep{Zhang2020SoftRigidHybrid}. Variable stiffness~\citep{Wolf2015VSAReview} and programmable actuation~\citep{Polygerinos2017HybridJoint} further boost adaptability.

Our prior work explores both soft and soft-rigid hands. The gripper in~\citep{Zhou2017SoftGripper} uses textured dual-chamber fingers for low-pressure adaptation. The BCL-26 hand~\citep{Zhou2019BCL26} achieves human-like motion with 26 fiber-reinforced soft degrees of freedom (DOFs). We incorporated stiffness tuning via particle jamming in~\citep{Zhou2020VSPP}. The DexCo Hand~\citep{Zhou2024DexCo} uses hydraulic actuation for forceful, controllable motion, while the Edgy-2 Hand~\citep{Zhou2018Edgy2} emphasizes modular pneumatic flexibility.

Despite recent advances, many studies remain focused on empirical evaluation and lack generalizable models for analyzing or comparing design variants. While task-specific tuning may suffice for grasping, manipulation tasks demand predictive insights into how structure and actuation affect performance. Two key metrics—\textit{manipulability}, describing motion capability~\citep{Doty1995Manipulability}, and \textit{compliance}, capturing passive deformation under force~\citep{Aukes2014UnderactuatedHand}—are essential for assessing dexterity and adaptability. Though common in rigid and underactuated systems, these metrics are rarely jointly analyzed in soft-rigid hybrid designs, especially across different actuation modes. This gap hinders structure-aware design and cross-system performance generalization.

To address this, we propose a unified analytical framework for evaluating both manipulability and compliance in modular soft-rigid hybrid fingers actuated by hydraulic or pneumatic systems. We formulate a Jacobian-based model that maps actuator inputs to joint and task-space behaviors. Hydraulic systems are modeled using a linear approximation under incompressible flow, while pneumatic actuators are captured using nonlinear pressure–volume relationships derived from ideal gas behavior~\citep{Polygerinos2017HybridJoint}. This formulation enables consistent computation of manipulability ellipsoids and compliance matrices across actuation types. We validate the proposed framework on two representative systems: the hydraulically actuated \textit{DexCo Hand} and the pneumatically actuated \textit{Edgy-2 Hand}. Simulation and experimental results reveal actuation-dependent trade-offs in directional dexterity and passive stiffness, offering insights for structure-aware analysis, actuator selection, and modular finger design in soft-rigid robotic fingers.

\begin{figure}
\centering
\begin{subfigure}{0.45\textwidth}
    \includegraphics[width=1\linewidth,]{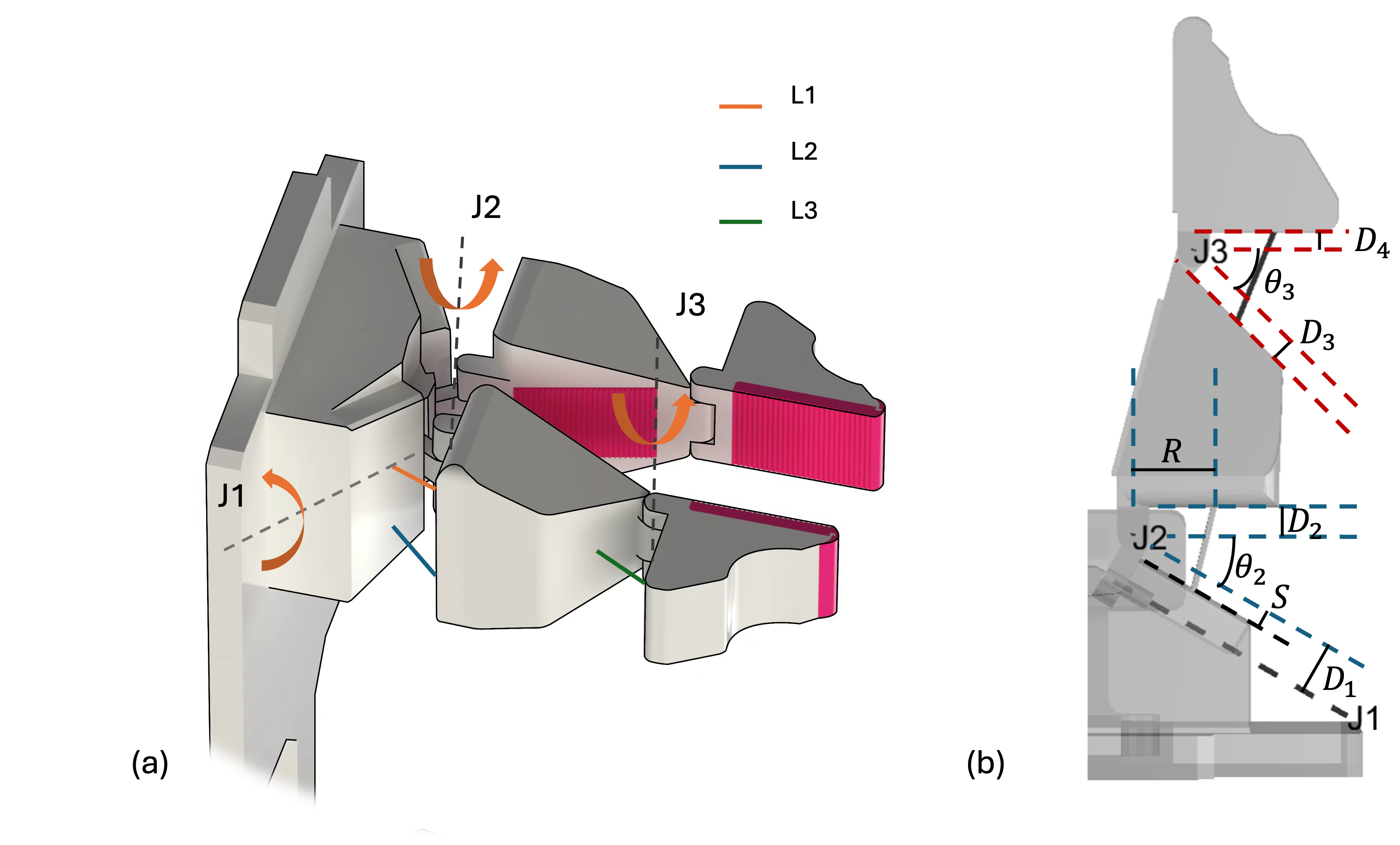}
\end{subfigure}
\caption{(a) Kinematic model of linear soft actuators: dashed lines (J1–J3) denote passive revolute joints; solid lines (L1–L3) represent soft actuators. (b) Top view of a DexCo finger at $\theta_1 = 0$, with parameters used in (\ref{eq:L1L2Maps1}) and (\ref{eq:L1L2Maps2}) labeled.}

\label{fig:PLabels}
\end{figure}

\section{Manipulability Analysis}
\label{sec:analysis}

To analyze the manipulability of the soft-rigid hybrid fingers, we begin with an approximate model that represents the soft actuators as soft linear actuators. Considering the inherent flexibility of soft actuators, we further assume that the soft linear actuators are connected to the knuckles of the DexCo hand via universal joints at both ends. Given the attachment points of the actuators in the local frame of each knuckle, and the poses of the revolute joints, the length of each linear soft actuator corresponds to the distance between its respective attachment points. A visualization of this model is provided in Fig. \ref{fig:PLabels}a.

This section analyzes the kinematic mapping between the lengths of the linear soft actuators, denoted as $l_1$, $l_2$ and $l_3$ and the poses of the passive revolute joints, denoted as $\theta_1$, $\theta_2$, $\theta_3$.  Based on visual inspection of Fig. \ref{fig:PLabels}, this mapping can be decomposed into two parts: $\theta_3$ depends solely on $l_3$ while $\theta_1$ and $\theta_2$ are jointly determined by $l_1$ and $l_2$. The remainder of this section discusses each of these relationships in detail and validates them via hardware experiments.

\subsection{Dual Soft Actuation Case}
\label{subsec:model-dual}

This subsection investigates the kinematic relationship between $l_1$, $l_2$ and the joint angles $\theta_1$, $\theta_2$. Through geometric analysis, this relationship can be expressed as follows:

\begin{equation}
\begin{split}
    l^2_1&=A^2+(H\cos{\theta_1}-V\sin{\theta_1}-S)^2 \\
         &+(H\sin{\theta_1}+V\cos{\theta_1}-V)^2 \\
    l^2_2&=A^2+(H\cos{\theta_1}+V\sin{\theta_1}-S)^2 \\
         &+(H\sin{\theta_1}-V\cos{\theta_1}+V)^2 \\
\end{split}
\label{eq:L1L2Maps1}
\end{equation}

Here, \(A\), \(H\), and \(V\) represent the axial, horizontal, and vertical offsets, respectively, from joint \(J_1\) to joint \(J_2\), as functions of \(\theta_2\) when \(\theta_1 = 0\). These offsets can be computed as follows. The parameter \(R\) denotes the radial distance from the attachment points of actuators \(L_1\) and \(L_2\) to the axis of joint \(J_2\). The quantities \(D_1\) and \(D_2\) represent the offsets between the actuator attachment surface and the rotation axes. The parameter \(S\) denotes the horizontal distance from joint \(J_1\) to the actuator base attachment point, as shown in Fig.~\ref{fig:PLabels}b. Lastly, \(L\) is the distance between the actuator endpoints on the attachment surface.

\begin{equation}
\begin{split}
    A&=R(1-\cos{\theta_2})+D_2\sin{\theta_1} \\
    H&=D_1+R\sin{\theta_2}+D_2\cos{\theta_2} \\
    V&=\frac{L}{2}
\end{split}
\label{eq:L1L2Maps2}
\end{equation}

\begin{figure}
\centering
    \includegraphics[width=0.75\linewidth]{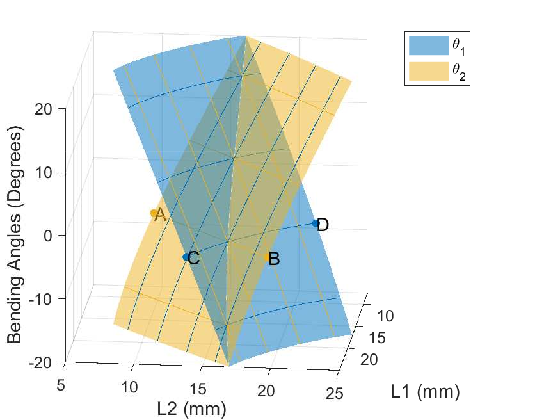}
\caption{Surface plot that maps $l1$ and $l2$ into $\theta_1$ and $\theta_2$. Yellow lines represent sets of $l_1$ and $l_2$ combinations where $\theta_1$ remains the same, while blues lines are sets where $\theta_2$ remains the same.}
\label{fig:t1t2_surface}
\end{figure}

Fig.~\ref{fig:t1t2_surface} illustrates the mapping between $(l_1, l_2)$ and $(\theta_1, \theta_2)$. We fix $\theta_1$ at $-15^\circ$, $-5^\circ$, $5^\circ$, and $15^\circ$, and plot the corresponding contour lines on the resulting surface. By vertically projecting these contour lines onto the $\theta_2$ surface, we obtain curves that represent the achievable range of $\theta_2$ through appropriate adjustment of $l_1$ and $l_2$. Similarly, another set of curves is generated to represent the achievable range of $\theta_1$ while holding $\theta_2$ fixed. The analysis is constrained to the range $[-20^\circ, 20^\circ]$ for both $\theta_1$ and $\theta_2$, as configurations beyond this range may result in self-collisions. The plots show that for each fixed value of $\theta_1$ or $\theta_2$, the other joint can still achieve the full range of $[-20^\circ, 20^\circ]$ through suitable adjustment of $l_1$ and $l_2$. This suggests that the pose of joint $J_1$ or $J_2$ has limited influence on the reachable workspace of the other joint.

\begin{figure}
\centering
\begin{subfigure}{0.24\textwidth}
    \includegraphics[width=\linewidth]{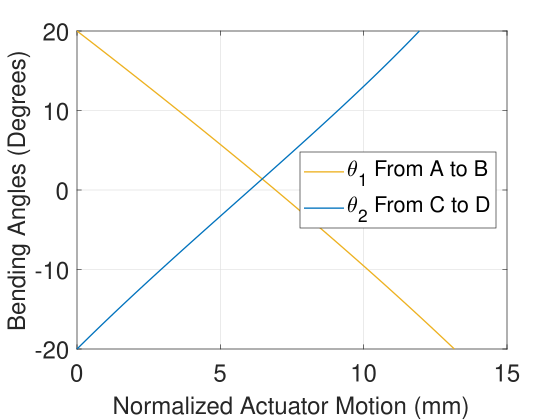}
    \subcaption{}
\end{subfigure}
\begin{subfigure}{0.24\textwidth}
    \includegraphics[width=\linewidth]{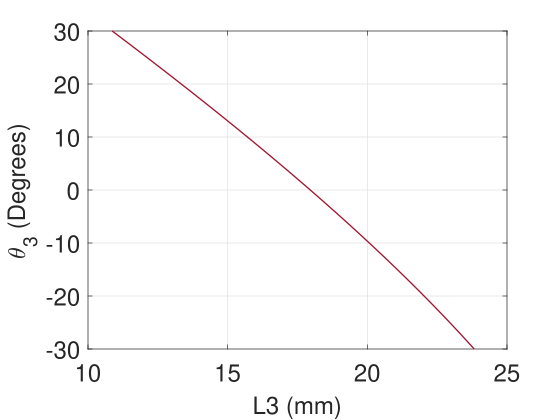}
    \subcaption{}
\end{subfigure}
\caption{(a): The projected curves for contours from A to B and from C to D. The positions of A, B, C, D are marked in Fig. \ref{fig:t1t2_surface}. (b): J3 pose against soft actuator L3 length.}
\label{fig:t1t2_fixed}
\end{figure}

To further analyze the relationship between actuator inputs and joint configurations, we plot the projected curves along the contour directions, as shown in Fig.~\ref{fig:t1t2_fixed}a. The two contour paths are defined from point A to B and from point C to D, as indicated in Fig.~\ref{fig:t1t2_surface}. The x-axis values in these plots correspond to a normalized actuator motion (denoted as $NAM$). Taking the contour from A to B as an example, for a given point $(l_1, l_2, \theta_1)$ on the contour, and letting $(l_{1,A}, l_{2,A}, \theta_1)$ denote the coordinates at point A, the normalized actuator motion is computed as follows:

\begin{equation}
    NAM=\sqrt{(l_1-l_{1,A})^2+(l_2-l_{2,A})^2}
\end{equation}

As shown in the left side of Fig.~\ref{fig:t1t2_fixed}, the motion of joints $J_1$ and $J_2$ exhibits an approximately linear relationship with respect to the normalized actuator motion, with a slope of approximately $3.5^\circ/\text{mm}$. This slope represents a balanced value, neither too steep nor too shallow, indicating that the soft actuators can effectively drive the passive joints without requiring excessive effort. At the same time, the passive joints are not overly sensitive to small variations in actuator motion, contributing to stable and controllable behavior.

\begin{figure}
\centering
    \includegraphics[width=0.7\linewidth]{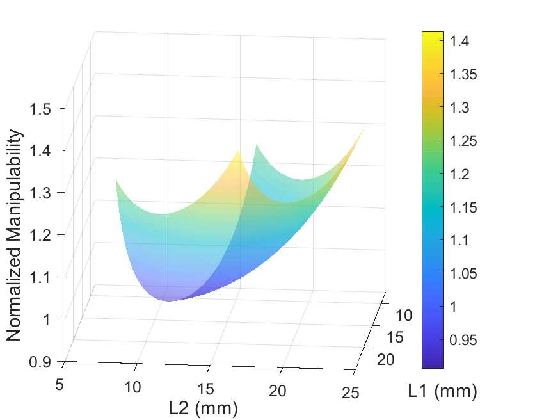}
\caption{Normalized manipulability distribution.}
\label{fig:t1t2_jac}
\end{figure}

Fig.~\ref{fig:t1t2_jac} presents the manipulability, $M$,  analysis over the entire workspace. The manipulability is computed using the following equation, which characterizes the actuator effort required to produce joint motions in arbitrary directions.

\begin{equation}
    M=\left|\det{\begin{bmatrix}
        \frac{\partial\theta_1}{\partial l_1} & \frac{\partial\theta_1}{\partial l_2} \\
        \frac{\partial\theta_2}{\partial l_1} & \frac{\partial\theta_2}{\partial l_2}
    \end{bmatrix}}\right|=\left(\left|\det{\begin{bmatrix}
        \frac{\partial l_1}{\partial\theta_1} & \frac{\partial l_1}{\partial\theta_2} \\
        \frac{\partial l_2}{\partial\theta_1} & \frac{\partial l_2}{\partial\theta_2}
    \end{bmatrix}}\right|\right)^{-1}
    \label{eq:jaco}
\end{equation}

The partial derivatives are directly obtained from (\ref{eq:L1L2Maps1}). To eliminate the influence of unit selection, the manipulability can be normalized by dividing it by its value at the reference configuration where $\theta_1 = \theta_2 = 0$. As shown in Fig.~\ref{fig:t1t2_jac}, the normalized manipulability remains well-conditioned across the workspace, without exhibiting any extreme or anomalous values.

\begin{figure}
\centering
    \includegraphics[width=0.7\linewidth]{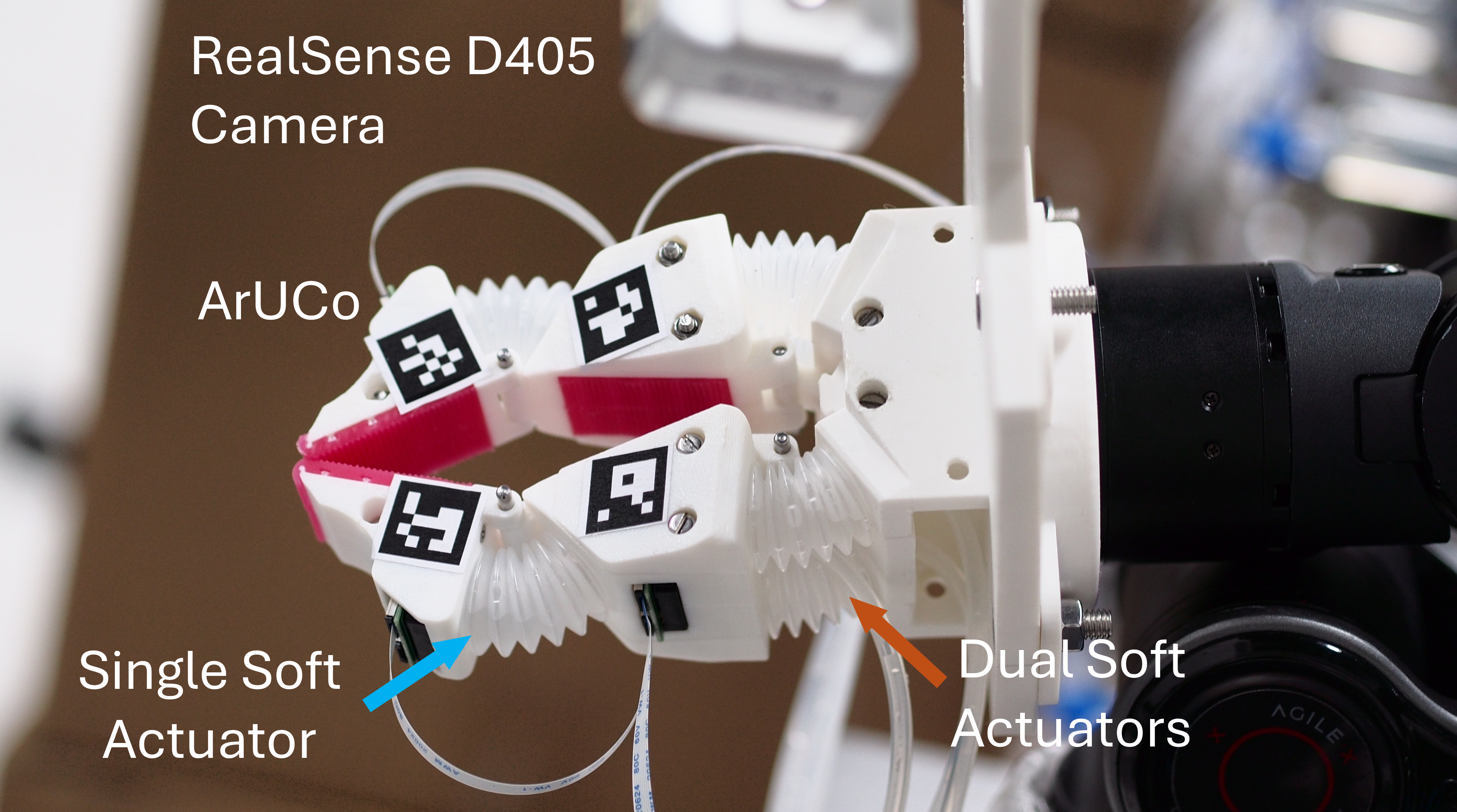}
\caption{Experimental setup. The joint angles are captured from ArUco code (\citep{romero2018speeded}).}
\label{fig:exp_setup}
\end{figure}

\begin{figure}
\centering
    \includegraphics[width=0.7\linewidth]{"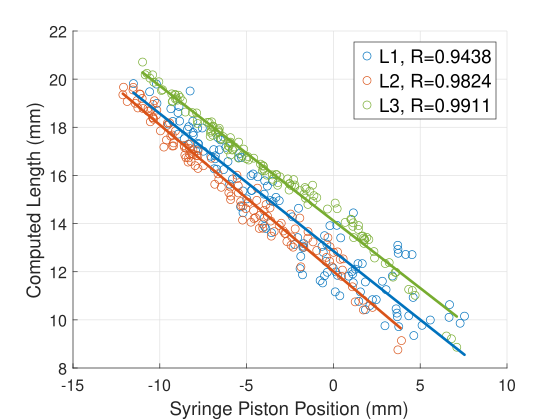"}
\caption{Linear regressions for hydraulic actuator lengths $l^e$ against syringe piston positions $Q^e$ in the experiments.}
\label{fig:exp_syringe}
\end{figure}

\subsection{Single Soft Actuation Case}
\label{subsec:model-single}

This section investigates the relationship between $l_3$ and $\theta_3$. Through direct analytical computation, the following expression can be derived:

\begin{equation}
\begin{split}
    l_3^2&=H_3^2+V_3^2 \\
    H_3&=D_3+R\sin{\theta_3}+D_4\cos{\theta_3} \\
    V_3&=R(1-\cos{\theta_3})+D_4\sin{\theta_3}
\end{split}
\label{eq:L3Maps}
\end{equation}

The parameters $D_3$ and $D_4$ represent the offsets between the soft actuator attachment surface and the rotational axes, as illustrated in Fig.~\ref{fig:PLabels}b. From this configuration, it can be observed in Fig.~\ref{fig:t1t2_fixed}b that $\theta_3$ exhibits an approximately linear relationship with $l_3$, with a slope of about $5^\circ/\text{mm}$. This suggests that joint $J_3$ responds to actuator $L_3$ with a balanced sensitivity, neither overly responsive nor excessively resistant to input, indicating effective and stable actuation.

\subsection{Experimental Validation with Hydraulic Actuators}

To validate the models proposed in (\ref{eq:L1L2Maps1}) and (\ref{eq:L3Maps}), we conducted experiments using a physical DexCo hand equipped with hydraulic actuators $L_1$, $L_2$, and $L_3$. These actuators are filled with incompressible liquid and actuated by three independent external syringes. Let $Q_1$, $Q_2$, and $Q_3$ denote the piston positions of the corresponding syringes. An increase in $Q$ indicates that liquid is being withdrawn from the actuator, resulting in a decrease in actuator length.

Due to the near-incompressibility of the liquid under room temperature and moderate pressure, the actuator lengths $l$ are expected to maintain an affine relationship with the piston positions, i.e., $l = kQ + b$, where $k$ and $b$ are scalar coefficients determined experimentally. Based on this assumption, we collected 200 data samples of the form $(Q_1^e, Q_2^e, Q_3^e, \theta_1^e, \theta_2^e, \theta_3^e)$, and computed the corresponding expected actuator lengths $l_1^e$, $l_2^e$, and $l_3^e$. Linear regression was then performed between the pairs $(Q_1^e, l_1^e)$, $(Q_2^e, l_2^e)$, and $(Q_3^e, l_3^e)$, as illustrated in Fig.~\ref{fig:exp_syringe}. The results demonstrate a strong linear correlation between actuator lengths and piston positions, with high coefficients of determination, validating the geometric models in (\ref{eq:L1L2Maps1}) and (\ref{eq:L3Maps}). The slightly higher variance observed in the L1 data may be attributed to minor misalignment or increased friction in its tubing connection, which affects the repeatability of piston-to-length translation in the physical setup.

\subsection{Potential Extension to Pneumatic Actuators}

It is worth noting that the geometric modeling presented in Sections~\ref{subsec:model-dual} and \ref{subsec:model-single} constitutes a generalizable framework that can also be applied to pneumatic actuators. While air is typically treated as a compressible fluid, it is common practice to establish a mapping from the piston position $Q$ and the pressure within the syringe $P$ to the actuator length $l$ through experimental calibration. Once this mapping is identified, it becomes possible to derive the relationship between $(Q, P)$ and the resulting joint configurations.

\begin{figure}
\centering
    \includegraphics[width=\linewidth]{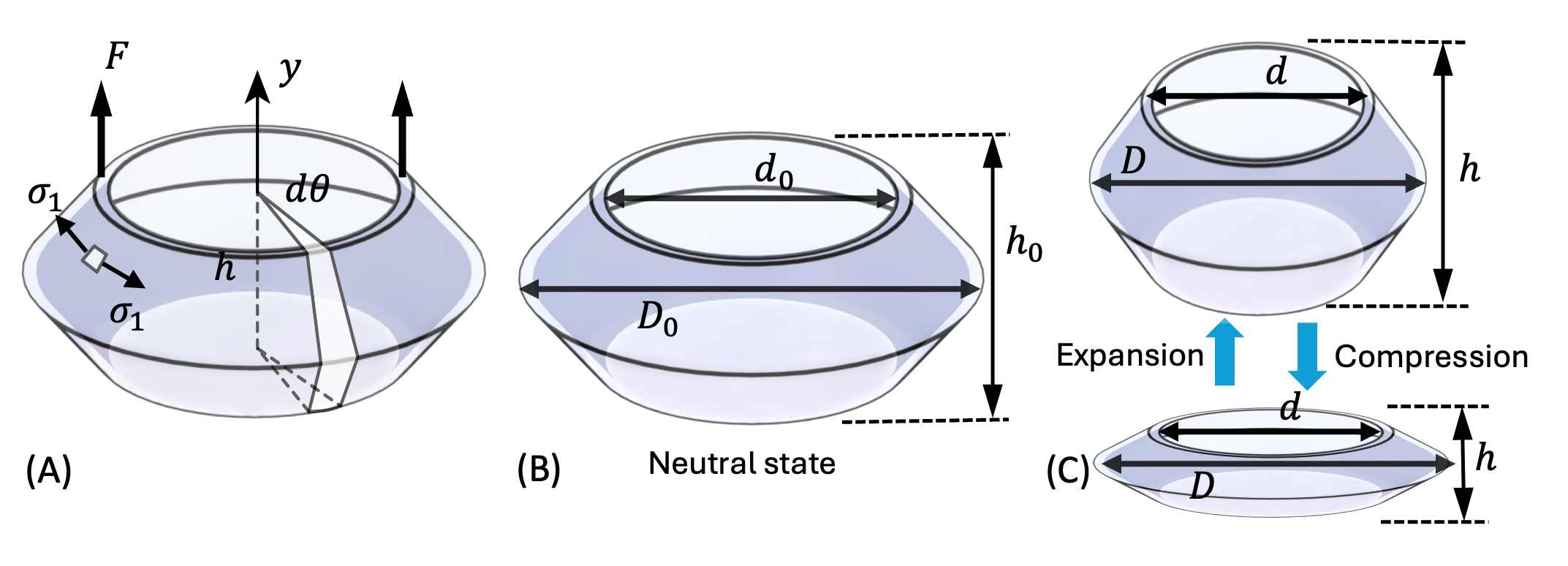}
\caption{(A) Compliance modeling for force estimation in actuation space. (B) Initial state of the origami bellow. (C) Extended and contracted states.}
\label{fig:force}
\end{figure}

\section{Compliance Modeling}

While manipulability characterizes motion capability, compliance captures the response to external forces—an equally critical aspect in soft-rigid design. Owing to material softness, both hydraulic and pneumatic actuators exhibit inherent compliance during interaction. This section presents a static analysis and compliance modeling of the soft-rigid hybrid hand, mapping actuation inputs ($\boldsymbol{a}$) to joint configurations ($\boldsymbol{\tau}$). We develop models for the compliance introduced by hydraulic and pneumatic actuation and validate them through object manipulation simulations.

\subsection{Hydraulic Compliance Modeling}

In hydraulic systems, compliance arises primarily from the compressibility of the working fluid, the elasticity of the fluidic chamber and tubing, and the mechanical compliance of the structure.

Compared to traditional hydraulic systems that exhibit high stiffness and low compliance, the dominant source of compliance in soft hydraulic actuators is the elastic deformation of the chamber material, characterized by its Young's modulus $E$. As the material deforms under pressure, the internal diameter changes from its undeformed size $D_0$ and $d_0$ to the deformed value $D$ and $d$ thereby altering the internal geometry and enabling displacement. The hydraulic force in soft actuators arises from the elastic deformation of the material in the axial direction during compression or extension, similar to inflating a balloon (Fig.~\ref{fig:force}).

We assume under an initial height $h_0$ of the soft actuator, the volume of water remains constant during motion, i.e., $V = V_0 \triangleq \text{constant}$. Based on this assumption, the geometric constraint is given by:
\begin{equation}
\begin{aligned}
V_0 &= \frac{\pi h_0}{3} \left[ \left(\frac{D_0}{2}\right)^2 + \left(\frac{d_0}{2}\right)^2 + \frac{D_0 d_0}{4} \right] \\
V &= \frac{\pi h}{3} \left[ \left(\frac{D}{2}\right)^2 + \left(\frac{d}{2}\right)^2 + \frac{D d}{4} \right] \\
V &= V_0
\end{aligned}
\label{eq:volume_constraint}
\end{equation}

To evaluate the radial deformation, we apply thin-walled pressure vessel theory. As illustrated in Fig. \ref{fig:force}, the hoop stress $\sigma_1$ is calculated along the actuator’s axial profile:

\begin{equation}
\sigma_1 = \frac{P r}{t}
\label{eq:hoop_stress}
\end{equation}

where $r$ is the local radius at height $y$, $P$ is the internal pressure, and $t$ is the wall thickness. According to the generalized Hooke’s law, the circumferential strain $\varepsilon_1$ is:

\begin{equation}
\varepsilon_1 = \frac{\sigma_1}{E} = \frac{\Delta_\ell}{2\pi r}
\label{eq:strain}
\end{equation}

where $\Delta_\ell$ denotes the change in circumference due to radial expansion, i.e., $\Delta_\ell = 2\pi (r - r_0)$, with $r_0$ representing the undeformed radius and $r$ the deformed radius. Using (\ref{eq:hoop_stress}) and (\ref{eq:strain}), we derive the expressions for the deformed inner and outer diameters under pressure $P$ via replacing $r$ with $\frac{d_0}{2}$ and $\frac{D_0}{2}$:

\begin{equation}
\begin{aligned}
d &= d_0 + \frac{2P}{E t} \left( \frac{d_0}{2} \right)^2 \\
D &= D_0 + \frac{2P}{E t} \left( \frac{D_0}{2} \right)^2
\end{aligned}
\label{eq:diameter_deformation}
\end{equation}

where the internal pressure is defined as $P \triangleq -\frac{4}{\pi d_0^2} F_h$. Substituting (\ref{eq:diameter_deformation}) into (\ref{eq:volume_constraint}), we can solve for the hydraulic force $F_h = F_h(h_0, \Delta y)$ by solving a resulting quadratic equation. 

To characterize the pressure-to-displacement relationship of the hydraulic actuator, we define two compliance-related parameters. The first is the hydraulic compliance $C_h$, which reflects the actuator’s deformation response to external force and is expressed as displacement per unit force. To translate this into a form compatible with pressure input modeling, we introduce the effective linear compliance $C_l = C_h / A$, where $A$ denotes the actuator’s effective cross-sectional area. This conversion yields a linear input-output relationship between internal pressure $P$ and output displacement $l$, given by $l = C_l \cdot P$. This formulation allows for consistent integration of actuator compliance into the Jacobian-based framework for joint-space stiffness and task-space compliance analysis.

In joint space, the Jacobian matrix $J \in \mathbb{R}^{n \times m}$ relates joint displacements $\Delta q \in \mathbb{R}^m$ to end-effector displacements $\Delta l \in \mathbb{R}^n$ via $\Delta l = J \cdot \Delta q$, as defined in Equation~(\ref{eq:jaco}). Under an external force $\Delta f \in \mathbb{R}^n$, the resulting torque deviation in joint space $\Delta \tau \in \mathbb{R}^m$ can be expressed as:

\begin{equation}
    \Delta \tau = J^T \cdot \Delta f = J^T \cdot \frac{1}{C_l} \cdot \Delta l = \frac{1}{C_l} J^T J \cdot \Delta q
\end{equation}

Here, $C_l$ is the effective linear compliance that maps input pressure to displacement. Based on this relationship, we derive the joint-space stiffness matrix $K_q \in \mathbb{R}^{m \times m}$ and its inverse (i.e., joint-space compliance matrix $C_q$) as:

\begin{equation}
    K_q = \frac{1}{C_l} J^T J, \quad C_q = K_q^{-1} = C_l \cdot (J^T J)^{-1}
\end{equation}

To analyze system behavior in task space, we transform these joint-space matrices using a task-space Jacobian $J_a \in \mathbb{R}^{n \times m}$, leading to:

\begin{equation}
    \begin{gathered}
    K_a = J_a^T K_q J_a \\
    J_a C_a J_a^T = C_q
    \end{gathered}
\end{equation}

where $K_a$ and $C_a$ are the task-space stiffness and compliance matrices, respectively. This transformation reveals how global actuator compliance $C_l$ interacts with configuration-dependent geometry (via $J$ and $J_a$) to shape the directional compliance profile in task space.

This result is analogous to classical manipulability-based compliance analysis, revealing how the hydraulic compliance $C_l$ globally scales the configuration-dependent compliance profile.

\subsection{Pneumatic Actuation Model}

In pneumatic systems, the input pressure is indirectly regulated via flow control, and the resulting output force is strongly dependent on chamber volume, which changes with deformation. Therefore the geometric constrain assumption made in hydraulic actuation is not ever applicable (\ref{eq:volume_constraint}).

Assuming quasi-static behavior and ideal gas behavior, we model the internal pressure as:

\begin{equation}
    P(l) = \frac{nRT}{V(l)} = \frac{nRT}{V_0 + \alpha l}
\end{equation}

where $l$ is the deformation-dependent variable (e.g., finger length or curvature), and $\alpha$ is a volume gain coefficient. The corresponding output force is:

\begin{equation}
    f(l) = \beta P(l) = \beta \cdot \frac{nRT}{V_0 + \alpha l}
\end{equation}

This nonlinear relation suggests a decreasing output force with increasing deformation, introducing variable stiffness properties in the actuation. When mapped to joint space via the Jacobian, the torque is:

\begin{equation}
    \tau = J^T(q) f(q)
\end{equation}

highlighting the pressure-volume-deformation coupling in the control of soft-rigid pneumatic systems. 

Experimental validation of pressure–deformation behavior in our previous work ~\citep{Zhou2018Edgy2} supports the pneumatic modeling in this work.

\subsection{Simulate Compliance Motion}

In simulation, we focus on simulating the hydraulic compliance. The variable stiffness property inherent in the DexCo hand can be modeled either as a linear variable elastic force, expressed as follows:

\begin{equation}
\tau_k = -k(\theta_0)\, \Delta \theta
\label{eq:linear_stiffness}
\end{equation}

where $\theta_0$ is the joint angle in the absence of external forces, $\theta$ is the current joint angle, and $\Delta 	\theta = 	\theta - \theta_0$ denotes the angular deformation. The linear elastic force in Eq.~\ref{eq:linear_stiffness} can be transformed into Cartesian space using the stiffness matrix $K_a = J_a^\top K_q J_a$, where $J_a$ is the Jacobian and $K_q$ is the joint stiffness matrix.

When modeling the stiffness as a torque input in simulation, it affects the system dynamics as follows:

\begin{equation}
M(\theta)\, \ddot{	\theta} + H(\dot{	\theta}, 	\theta) + G = \tau_k
\label{eq:dynamics_with_stiffness}
\end{equation}

\begin{figure}
\centering
    \includegraphics[width=1\linewidth]{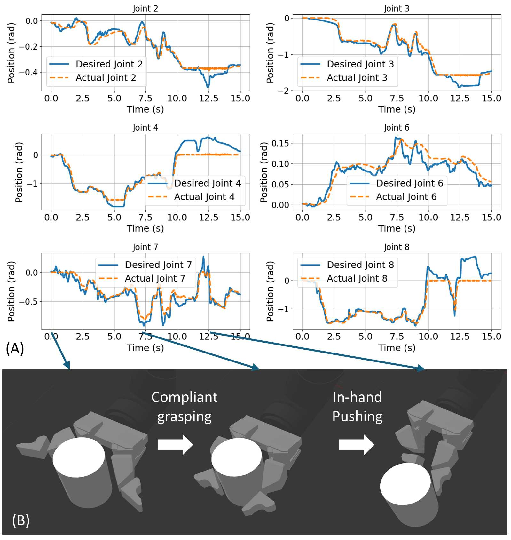}
\caption{Compliant motion simulation of the Soft-Rigid hybrid hand in Gazebo: (A) joint trajectories; (B) final interaction state. }
\label{fig:sim}
\end{figure}

In Eq.~\ref{eq:dynamics_with_stiffness}, the left-hand side captures inertial, Coriolis, and gravitational effects computed by the Gazebo simulator. The right-hand term, $\tau_k$, represents torques from elastic deformation. Incorporating $\tau_k$ enables simulation of nonlinear variable stiffness behaviors, as shown in Fig.~\ref{fig:sim}B.

The simulation involves two phases: a compliant caging grasp to enclose the object, followed by forceful release via increased base joint actuation and fingertip reorientation. Fig.~\ref{fig:sim}A shows both commanded and actual joint angles. Compliance-induced deviations are observed between target and actual responses, where Joints 2 and 3 form the base, and Joint 4 corresponds to the fingertip.

\section{Conclusion}
This paper presents a unified framework for analyzing the manipulability and compliance of modular soft-rigid hybrid fingers actuated by hydraulic and pneumatic systems. Using a generalized Jacobian-based formulation, we enable consistent structural evaluation of directional dexterity and passive compliance across actuation modes. Experimental validation on the DexCo (hydraulic) and Edgy-2 (pneumatic) hands highlights key trade-offs between actuation type, structure, and control sensitivity.

Our results highlight that control response and actuation speed are distinct in hybrid systems. As shown in Table 1, hydraulic actuators enable precise control but slower motion, while pneumatic actuators offer faster deformation with less stability. These mismatches suggest the need for functional separation or hierarchical integration in control design.

\begin{table}[h]
\setlength{\abovecaptionskip}{2pt}     
\setlength{\belowcaptionskip}{-5pt}
\captionsetup{font=scriptsize}  
\caption{Summary of actuation trade-offs.}
\centering
\scriptsize
\begin{tabular}{|l|c|c|}
\hline
\textbf{Metric} & \textbf{Hydraulic} & \textbf{Pneumatic} \\
\hline
Control response speed & Fast & Slow \\
Actuation speed & Slower & Faster \\
Force output precision & High & Low--moderate \\
Passive compliance & Low & High \\
\hline
\end{tabular}
\end{table}

We recommend structure-aware and actuation-aware design principles, including: (1) avoiding direct coupling of mismatched subsystems; (2) assigning distinct roles to each actuator type (e.g., pneumatic for grasping, hydraulic for in-hand manipulation); and (3) leveraging our framework to assess design trade-offs early in development.

While the current formulation focuses on quasi-static behavior, future work will incorporate dynamic effects such as inertia and damping, particularly for soft-rigid hybrid fingers. We also aim to extend the framework to include contact interactions and grasp stability for task-level analysis and robust control.

\label{secLconclusion}

\bibliography{main}                                                          % with bibtex (preferred)
                                                   
%\begin{thebibliography}{xx}  % you can also add the bibliography by hand

%\bibitem[Able(1956)]{Abl:56}
%B.C. Able.
%\newblock Nucleic acid content of microscope.
%\newblock \emph{Nature}, 135:\penalty0 7--9, 1956.

%\bibitem[Able et~al.(1954)Able, Tagg, and Rush]{AbTaRu:54}
%B.C. Able, R.A. Tagg, and M.~Rush.
%\newblock Enzyme-catalyzed cellular transanimations.
%\newblock In A.F. Round, editor, \emph{Advances in Enzymology}, volume~2, pages
%  125--247. Academic Press, New York, 3rd edition, 1954.

%\bibitem[Keohane(1958)]{Keo:58}
%\newblock \emph{Power and Interdependence: World Politics in Transitions}.
%\newblock Little, Brown \& Co., Boston, 1958.

%\bibitem[Powers(1985)]{Pow:85}
%\newblock Is there a way out?
%\newblock \emph{Harpers}, pages 35--47, June 1985.

%\bibitem[Soukhanov(1992)]{Heritage:92}
%\newblock \emph{{The American Heritage. Dictionary of the American Language}}.
%\newblock Houghton Mifflin Company, 1992.

%\end{thebibliography}

% \appendix
% \section{A summary of Latin grammar}    % Each appendix must have a short title.
% \section{Some Latin vocabulary}              % Sections and subsections are supported  
                                                                         % in the appendices.
\end{document}